\documentclass[runningheads]{llncs}
\usepackage{graphicx}
\usepackage{booktabs}
\usepackage{tabularx}
\usepackage{multirow}
\usepackage{rotating}
\usepackage{amsmath}
\usepackage{tikz}
\usepackage{verbatim}

\usetikzlibrary{arrows,arrows.meta,shapes}

\usepackage[caption=true,font=footnotesize]{subfig}
\begin{document}
	\pgfdeclarelayer{background}
	\pgfsetlayers{background,main}
	\title{Can Genetic Programming Do\\ Manifold Learning Too?}
	%
	%
	\author{Andrew Lensen\orcidID{0000-0003-1269-4751} \and
	Bing Xue \and
	Mengjie Zhang}%
	%
	%
	\institute{School of Engineering and Computer Science,\\
		Victoria University of Wellington, PO Box 600, Wellington 6140, New Zealand\\
	\email{\{andrew.lensen,bing.xue,mengjie.zhang\}@ecs.vuw.ac.nz}}
	\maketitle              
	\vspace{-1em}
	\begin{abstract}
	Exploratory data analysis is a fundamental aspect of knowledge discovery that aims to find the main characteristics of a dataset. Dimensionality reduction, such as manifold learning, is often used to reduce the number of features in a dataset to a manageable level for human interpretation. Despite this, most manifold learning techniques do not explain anything about the original features nor the true characteristics of a dataset. In this paper, we propose a genetic programming approach to manifold learning called GP-MaL which evolves functional \textbf{mappings} from a high-dimensional space to a lower dimensional space through the use of interpretable trees. We show that GP-MaL is competitive with existing manifold learning algorithms, while producing models that can be interpreted and re-used on unseen data. A number of promising future directions of research are found in the process.
		
	\end{abstract}
	\section{Introduction}
	\label{sec:introduction}
	Manifold learning has risen to prominence in recent years due to significant improvement in autoencoders and the widespread use of the t-Distributed Stochastic Neighbour Embedding (t-SNE) visualisation algorithm \cite{maatenTSNE}. Manifold learning is the main area in the non-linear dimensionality reduction literature, and consists of algorithms which seek to discover an embedded (non-linear) manifold within a high-dimensional space so that the manifold can be represented in a much lower-dimensional space. Hence, they aim to perform dimensionality reduction while preserving as much of the \textit{structure} of the high-dimensional space.
	
	Within manifold learning, there are two broad categories of algorithms: those that produce a mapping between the high and low-dimensional spaces (``mapping methods''), and those which provide only the found embedding\footnote{An embedding here refers to the low-dimensional representation of the structure present in a dataset.} (``embedding methods''). Mapping methods are particularly attractive, as they allow future examples to be processed without re-running the algorithm, and they have the potential to be \textit{interpretable}, which is often desirable in machine learning tasks.
	
	Genetic Programming (GP) is well known for producing \textit{functions} which map inputs (the domain) to outputs (the codomain) using tree-based GP \cite{poli2008field}. GP appears to have several promising characteristics for solving this problem: 
	
	\begin{itemize}
		\item It is a global learner, and so should be less prone to producing partial-manifolds (i.e.\ local minima) unlike many existing methods which use gradient descent or other approaches;
		\item As it uses a population-based search method, it does not require a differentiable fitness function (unlike auto-encoders, t-SNE, etc.) and so could be used with a range of optimisation criteria; and
		\item It is intrinsically suited to producing interpretable mappings, as tree-based GP in particular can be understood by evaluating the tree from bottom to top. A wide range of tools are also available for producing interpretable GP models, including automatic program simplification and parsimony pressure.
	\end{itemize}
	Despite these traits, we are not aware of any work that uses GP to learn a manifold by mapping an input dataset to a set of lower-dimensional outputs.
	
	\subsection{Goals}
	\label{sec:goals}
	In this work, we propose the first approach to using GP to perform Manifold Learning (\textbf{GP-MaL}). In particular, we will:
	
	\begin{enumerate}
		\item Propose a multi-tree GP representation and function and terminal sets for performing manifold learning;
		\item Construct an appropriate fitness function to evaluate how effectively a GP individual preserves the structure of the high-dimensional space;
		\item Evaluate how GP-MaL fares compared to existing manifold learning algorithms on a variety of classification tasks; and
		\item Investigate the viability of GP-MaL for producing interpretable mappings of manifolds.
	\end{enumerate}

	\section{Background}
	\label{sec:background}
	\subsection{Dimensionality Reduction}
	Broadly speaking, dimensionality reduction (DR) is the task of reducing an existing feature space into a lower-dimensional one, which can be better understood and processed more efficiently and effectively \cite{liu2012feature}. Two main approaches in DR are feature selection (FS) and feature extraction/construction \cite{liu2012feature}. While FS approaches --- which select a small subset of the original features --- are sufficient when a dataset has significant intrinsic redundancy/irrelevancy, there is a limit to how much the dimensionality can be reduced by FS alone. For example, if we want to reduce the dimensionality to two or three features, using FS alone is likely to poorly retain the structure of the dataset. In such a scenario, feature extraction/construction methods are able to reduce dimensionality more effectively by \textit{combining} aspects of the original features in some manner.
	
	One of the most well-known FC methods is Principle Component Analysis (PCA) \cite{jolliffe2011pca}. PCA produces \textit{components} (constructed features) which are linear combinations of the original features, such that each successive component has the largest variance possible while being orthogonal to the preceding components. Variance is a fundamental measurement of the amount of \textit{information} in a feature, and so PCA is optimal for performing linear dimensionality reduction under this framework. However, linear combinations are not sufficient when data has a complex underlying structure; linear methods tend to focus on maintaining global structure while struggling to maintain local neighbourhood structure in the constructed feature space \cite{maatenTSNE}. Thus, there is a clear need for nonlinear dimensionality reduction, of which a major research area is manifold learning.
	
	\subsection{Manifold Learning}
	\label{sec:manifoldLearning}
	Manifold learning algorithms are based on the assumption that the majority of real-world datasets have an intrinsic redundancy in how they represent information they contain through their features. A manifold is the inherent underlying structure which contains the information held within that dataset, and often this manifold can be represented using a smaller number of features than that of the original feature set \cite{bengio2013representation}. Thus, manifold learning algorithms attempt to learn/extract this manifold into a lower-dimensional space. PCA, for example, can be seen as a linear manifold learning algorithm; of course, most real-world manifolds are strongly non-linear \cite{bengio2013representation}. 
	
	Multidimensional Scaling (MDS) \cite{kruskal1964mds} was one of the first approaches to manifold learning proposed, and attempts to maintain between-instance distances as well as possible from the high to the low dimensional space. Metric MDS often uses a loss function called \textit{stress}, which is then minimised using a majorizing function from convex analysis. Another well-known, more recent method is Locally-Linear Embedding (LLE) \cite{roweis2000lle}, which describes each instance as a linear combination of its neighbours \footnote{Here, neighbours refer to the closest instances to a point by (Euclidean) distance.}, and then seeks to maintain this combination in the low-dimensional space using eigenvector-based optimisation. MDS performs a non-parametric transformation of the original feature space, and so is not interpretable with respect to the original features; LLE is also difficult to interpret given it is based on preserving neighbourhoods.
	
	t-Distributed Stochastic Neighbor Embedding (t-SNE) \cite{maatenTSNE} is considered by many to be the state-of-the-art method for performing visualisation (i.e.\ 2D/3D manifold learning); it models the original feature space as a joint probability distribution in terms of how close an instances' neighbours are and then attempts to produce the same joint distribution in the low-dimensional space by using Kullback–Leibler divergence to measure the similarity of the two distributions. However, t-SNE was developed purely for visualisation (2/3D dimensionality reduction) and so it is not specifically designed as a general manifold learning algorithm \cite{maatenTSNE}. It is also similar to MDS in that it produces an embedding with no mapping back to the original features. Finally, autoencoders are often regarded to do a type of manifold learning \cite{bengio2013representation}, but again they tend to be quite opaque in the meaning of their learnt representation, while requiring significantly more computational resources than the classical manifold learning methods.

	\subsection{Related Work}
	\label{sec:relatedWork}
	Evolutionary Computation (EC) has seen very recent use in evolving autoencoders for image classification tasks using Genetic Algorithms \cite{sun2018evolving}, GP \cite{rodriguez2018structurally}, and Particle Swarm Optimisation \cite{sun2018pso}. Historically, auto-encoders have had to be manually designed or require significant domain knowledge to get good results, and so automatic evolution of auto-encoder structure is a clear improvement. However, these methods are still a somewhat indirect use of EC for representation/manifold learning, as they do not allow an EC method to directly learn the underlying structure as our proposed GP approach may.

	GP has also been used for visualisation (i.e.\ the 2D form of manifold learning) in a supervised learning context using a multi-objective fitness function to optimise both classification performance and clustering-based class separability measures \cite{cano2017multi}. Recently, a GP method was proposed to evolve features for feature selection algorithm testing \cite{lensen2018automatically}, which also used a multi-tree representation, but used a specialised fitness function to encourage redundant feature creation based on mutual information (MI). The use of GP for visualisation of solutions for production scheduling problems has also been recently investigated \cite{nguyen2018visualising}. GP has also been applied to other tangential unsupervised learning tasks for feature creation, such as clustering \cite{lensen2017New}, as well as extensive use in supervised learning domains \cite{tran2016genetic,neshatian2012filter}. Clearly, GP has shown significant potential as a feature construction method, and so it is hoped that it can be extended to directly perform manifold learning as well.

	\section{GP for Manifold Learning (GP-MaL)}
	The proposed method, GP-MaL, will be introducted in three stages. Firstly, the design of the terminal and function set is discussed. Then, a fitness function appropriate for manifold learning is formulated and explained. Finally, a method to improve the computational efficiency of GP-MaL (while maintaining good performance) is developed.
	\subsection{GP Representation}
	In this work, we utilise a multi-tree GP representation, whereby each tree represents a single dimension in the output (low-dimensional) space. While multi-tree GP is known to scale poorly as the number of trees ($t$) increases, manifold learning usually assumes a low output dimensionality (e.g.\ $t < 10$). The terminal set consists of the $d$ scaled real-valued input features, as well as random constants drawn from $U[-1,+1]$ to allow for variable sub-tree weighting. The output of each tree is not scaled or normalised in any way as this may introduce bias to the evolved trees or affect tree interpretability.
	
	The function set (Table \ref{functionSet}) chosen is inspired by existing feature construction and manifold learning literature. It includes the standard ``$+$'' and ``$\times$'' arithmetic operators to allow simple combinations of features/sub-trees, as well as a ``$5+$'' operator which sums over five inputs\footnote{Five inputs were found to be a good balance between encouraging wider trees and minimising computing resources required.} to encourage the use of many input features on large datasets. Subtraction and division were not included as they are the complements of addition and subtraction and so are redundant in the ``way'' in which they combine sub-trees. To encourage the learning of non-linear manifolds, two common non-linear activation functions from auto-encoders were added: the sigmoid and rectified linear unit (ReLU) operators. The function set also includes two conditional (non-differentiable!) operators, ``max'' and ``min'', which may allow GP to produce more advanced functions. Finally, the ``if'' function is also included, which takes three inputs $a, b, c$ and outputs $b$ if $a > 0$ or $c$ otherwise, to allow for more flexible conditions to be learnt.

\begin{table}[t]
	\centering
	\vspace{-1.25em}
	\caption{Summary of the function set used in GP-MaL.}
	\label{functionSet}
		\begin{tabularx}{\textwidth}{@{}lXlllXccXlll@{}}
			\toprule
			Category  & & \multicolumn{3}{c}{Arithmetic} && \multicolumn{2}{c}{Non-Linear} && \multicolumn{3}{c}{Conditional} \\ \cmidrule{3-5}\cmidrule{7-8}\cmidrule{10-12} 
			Function &  & $+$\hspace{1em}    & $\times$\hspace{1em}    & $5+$ &    & Sigmoid   \hspace{1em}      & ReLU      &   & Max    \hspace{1em}   & Min  \hspace{1em}     & If      \\ 
			
			No.\ of Inputs & & 2      & 2           & 5    &   & 1               & 1        &    & 2         & 2         & 3       \\
			No.\ of Outputs && 1      & 1           & 1     &  & 1               & 1        &    & 1         & 1         & 1       \\ \bottomrule
		\end{tabularx}%
\vspace{-1.5em}
\end{table}

Mutation is performed by selecting a random tree in a GP individual, and then selecting a random sub-tree to mutate within that tree, as standard. Crossover is performed in a similar way, by selecting a random tree from each candidate individual, and then performing standard crossover. 

	\subsection{Fitness Function}
	A common optimisation strategy among manifold learning algorithms is to encourage preserving the high-dimensional neighbourhood around each instance in the low-dimensional space. For example, MDS attempts to maintain distances between points, whereas t-SNE uses a probabilistic approach to model how related different points are, and attempts to produce an embedding with a similar joint probability distribution. We refrain from using a distance-based approach due to the associated issues with the curse of dimensionality \cite{francois2007concentration}, and instead try to preserve the \textit{ordering} of neighbours from the high to low dimensions.
	
	Consider an instance $I$ which has ordered neighbours $N=\{N_1,N_2,...,N_{n-1}\}$ for $n$ instances neighbours in the high-dimensional space, and neighbours $N'$ in the low-dimensional space. If we were to perfectly retain all structure in the dataset, then the ordering of $N'$ must be identical to that of $N$, i.e. $N = N'$. In other words, the quality of the low-dimensional space can be measured by how \textit{similar} $N'$ is to $N$. In this work, we propose measuring similarity by how far each instances' neighbours deviate in their ordering in the low-dimensional space compared to the high-dimensional space. For example, if $N=\{N_1,N_2,N_3\}$ and $N'=\{N_2,N_3,N_1\}$, the neighbours deviate by 2, 1, and 1 positions respectively. Clearly, the larger the deviation, the more inaccurately the orderings have been retained. Let $Pos(a,X)$ give the index of $a$ in the ordering of $X$. We propose the following similarity measure:
	\begin{equation}
	Similarity(N, N') = \sum_{a \in N} Agreement(|Pos(a,N) - Pos(a,N')|)
	\end{equation}
	where $Agreement$ is a function that gives \textbf{higher} values for \textbf{smaller} deviations. GP-MaL uses an $Agreement$ function based on a Gaussian weighting to allow for small deviations without significant penalty, while still penalising large deviations harshly. In this work, a Gaussian with a $\mu$ of $0$ and $\theta =20$ is used. $\theta$ controls how harshly deviations are punished -- in preliminary testing we found a high $\theta$ gave best results as it created a smoother fitness landscape. The weighting for a given deviation $dev$ is $1 - prob(-dev,+dev)$, i.e. the area of the Gaussian not in this range. In this way, when there is no deviation, the weighting is $1$ (perfect), whereas when it is maximally deviated the weighting tends to $0$.
	
	The complete fitness function is the normalised similarity across all instances in the dataset ($X$):
			\vspace{-.5em}
	\begin{equation}
	Fitness = \frac{1}{n^2}\sum_{I \in X} Similarity(N_I, N^{'}_{I})
			\vspace{-.5em}
	\end{equation}
	Fitness is in the range $[0,1]$ and should be maximised. 
	\subsection{Tackling the Computational Complexity}
	Unfortunately, computing the above fitness requires ordering every instance's neighbours by their distances in the low-dimensional space, at a cost of $O(n \log(n)$ using a comparison sort. This gives a net complexity of $O(n^2 \log(n))$ for each individual in the population. This scales poorly with the number of instances in the dataset. Consider a given neighbour $N_b$, which comes after $N_a$ and before $N_c$. Even if we do not optimise the deviation of $N_b$, it seems likely that it will still be near $N_a$ and $N_c$ in the low-dimensional ordering, as it is likely to have similar feature values to $N_a$ and $N_c$ and hence will have a similar output from the evolved function. Based on this observation, we can omit some neighbours from our similarity function in order to reduce the computational complexity. Clearly, removing any neighbours will slightly reduce the accuracy of the fitness function, but this is made up by the significantly decreased computational cost (similar to surrogate model approaches). An example of this can be seen in Fig.\ \ref{graphFig}, where the number of edges are decreased significantly by only considering the two nearest neighbours. Despite this, the global structure of the graph is still preserved well, with E and G only having two edges to the rest of the (distant) nodes, and C, D, and F sharing many edges as they are in close proximity.

	\tikzset{>=stealth}
	
	\begin{figure}[b!]
		\vspace{-2em}
	\captionsetup[subfigure]{width=.49\textwidth}
\	
\subfloat[Complete graph: 42 directed edges.]{
	
	\resizebox{.3\textwidth}{!}{%
	
\begin{tikzpicture}
\begin{scope}[every node/.style={circle,thick,draw}]
\node (A) at (0,1) {A};
\node (B) at (0,3) {B};
\node (C) at (2.5,4) {C};
\node (D) at (3.5,1) {D};
\node (E) at (2.5,-2.5) {E};
\node (F) at (4.5,3) {F} ;
\node (G) at (3.5,-2.5) {G};
\end{scope}

\begin{scope}[
every node/.style={fill=white,circle},
every edge/.style={draw=black,thick}]
\path [<->] (A) edge (B);
\path [<->] (A) edge (C);
\path [<->] (A) edge (D);
\path [<->] (A) edge (E);
\path [<->] (A) edge (F);
\path [<->] (A) edge (G);
\path [<->] (B) edge (C);
\path [<->] (B) edge (D);
\path [<->] (B) edge (E); 
\path [<->] (B) edge (F);
\path [<->] (B) edge (G);
\path [<->] (C) edge (D);
\path [<->] (C) edge (E);
\path [<->] (C) edge (F);
\path [<->] (C) edge (G);
\path [<->] (D) edge (E);
\path [<->] (D) edge (F);
\path [<->] (D) edge (G);
\path [<->] (E) edge (F); 
\path [<->] (E) edge (G);
\path [<->] (F) edge (G);

\end{scope}
\end{tikzpicture}
}
}
\hfil
\subfloat[Pruned graph: each node is connected to its two nearest neighbours. 14 edges.]{
		\resizebox{.3\textwidth}{!}{%
\begin{tikzpicture}
\begin{scope}[every node/.style={circle,thick,draw}]
\node (A) at (0,1) {A};
\node (B) at (0,3) {B};
\node (C) at (2.5,4) {C};
\node (D) at (3.5,1) {D};
\node (E) at (2.5,-2.5) {E};
\node (F) at (4.5,3) {F} ;
\node (G) at (3.5,-2.5) {G};
\end{scope}

\begin{scope}[
every node/.style={fill=white,circle},
every edge/.style={draw=red,very thick}]
\path [->] (A) edge (B);
\path [->] (A) edge (D);
\path [->] (B) edge (A);
\path [->] (B) edge (C);
\path [->] (C) edge (B);
\path [->] (C) edge (F);
\path [->] (D) edge (C);
\path [->] (D) edge (F);
\path [->] (E) edge (A); 
\path [->] (E) edge (G);
\path [->] (F) edge (C);  
\path [->] (F) edge (D);  
\path [->] (G) edge (D);  
\path [->] (G) edge (E);  

\end{scope}
\end{tikzpicture}
}%
}
\caption{Pruning of a graph to reduce computational complexity.}
\label{graphFig}
\vspace{-1.5em}
	\end{figure}
	
When considering which neighbours to omit, it is more important to consider the closer neighbours' deviations, in order to preserve local structure, as this is most likely to preserve useful information about relationships in the data. However, it is still important to consider more distant neighbours, so that the global structure is also preserved. Based on this, we propose choosing neighbours more infrequently the further down the nearest-neighbour list they are. One approach is to choose the first $k$ neighbours, followed by $k$ of the next $2k$ neighbours (evenly spaced), then $k$ of the next $4k$, etc. This gives $\eta$ neighbours according to the following equation:
\vspace{-.25em}
	\begin{equation}
	\eta = k \log_2({\frac{n}{k}}+1)
	\end{equation}
	thus $\eta$ is proportional to $\log(n)$ ($k \ll n$). The complexity per GP individual is then $O(\eta \log(\eta)) = O(\log(n) \log(\log(n)))$,  which gives a sublinear complexity. 
	In preliminary testing, we found using $k=10$ to give only minor differences in learning performance which was significantly outweighed by the ability to train for many more generations in the same computational time. We use this approach in all experiments in this paper. While $k$ could be perhaps be decreased further, it would not reduce computational time significantly, as tree evaluation is now the main cost of the evolutionary process.

	\section{Experiment Design}
	To evaluate the quality of our proposed GP-MaL algorithm, we focus mainly on the attainable accuracy on classification datasets using the evolved low-dimensional datasets. High classification accuracy generally requires as much of the structure of the dataset to be retained as possible in order to find the best decision boundaries between classes, and so is a useful proxy for measuring the amount of retained structure. We refrain from using the fitness function (or similar optimisation criteria) to measure the manifold ``quality'' so as not to introduce bias towards any specific manifold learning method. The scikit-learn \cite{scikit-learn} implementation of the Random Forest (RF) classification algorithm (with 100 trees) is used as it is a widely used algorithm with high classification accuracy, is stable across a range of datasets, and has reasonably low computational cost \cite{zhang2017comparison}. While other algorithms could also be compared, we found the results to be generally consistent across algorithms, and so do not include these for brevity. 10-fold cross-validation is used to evaluate every generated low-dimensional dataset, and 40 evolved datasets (40 GP runs) are used for each tested dataset in order to account for evolutionary stochasticity.
	
	The characteristics of the ten datasets we used for our experiments are shown in Table \ref{table:datasets}. A range of datasets from varying domains were chosen with different numbers of features, instances, and classes. 
	

		\renewcommand{\arraystretch}{1.1}
\begin{table}[t]
\vspace{-2em}
	\caption{Classification datasets used for experiments. Most datasets are sourced from the UCI repository \cite{uci} which contains original accreditations.}
		\label{table:datasets}
	\begin{tabularx}{\textwidth}{@{}lrrrXlrrr@{}}
		\toprule
		Dataset & Instances & Features & Classes &  & Dataset & Instances & Features & Classes \\ \cmidrule{1-4} \cmidrule{6-9}

Wine & 178 & 13 & 3 &  & COIL20 & 1440 & 1024 & 20 \\
Movement Libras & 360 & 90 & 15 &  & Madelon & 2600 & 500 & 10 \\
Dermatology & 358 & 34 & 6 &  & Yale & 165 & 1024 & 15 \\
Ionosphere & 351 & 34 & 2 &  & MFAT & 2000 & 649 & 10 \\
\begin{tabular}[c]{@{}l@{}}Image \\ Segmentation\end{tabular} & 2310 & 19 & 7 & \multicolumn{1}{l}{} & \begin{tabular}[c]{@{}l@{}}MNIST\\ 2-class\end{tabular} & 2000 & 784 & 2\\
\bottomrule

	\end{tabularx}
\vspace{-1.5em}
\end{table}

	We compare the proposed GP-MaL method to a number of baseline manifold learning methods: PCA (as a linear baseline), MDS (which uses a similar optimisation criteria), LLE (a popular MaL method) and t-SNE (state of the art for 2D/3D manifold learning). Scikit-learn \cite{scikit-learn} was used for all the baseline methods, except for t-SNE, with default settings. For t-SNE, we used van der Maaten's more efficient Barnes-Hut implementation \cite{maaten2014accelerating}. For each method and dataset, we produce transformed datasets for two, three, five, and the cube root (CR) of the number of original features. Two/three features are useful for visualisation but are unlikely to be sufficient to preserve all structure, whereas the cube root approach was found in preliminary testing to be the point at which all tested methods could capture maximal structure from the datasets. Five features are used as a ``middle-ground''. As all of these implementations have stochastic components, we also ran each 40 times for each dataset.
	
	We use standard GP parameter settings, as per Table \ref{table:parameterSettings}. One notable setting is that we use 1000 generations; as we are interested primarily in exploring the \textit{potential} of GP for this task, we are not particularly concerned with optimising the number of generations for best efficiency; this will be explored in future work. 
	
	\begin{table}[t]
		
		\centering
		\vspace{-.5em}
		\caption{GP Parameter Settings.}
		\label{table:parameterSettings}
		\begin{tabularx}{0.75\textwidth}{ll X ll}
			
			\toprule
			Parameter& Setting && Parameter & Setting\\
			\cmidrule(r){1-2}  \cmidrule(l){4-5}
			Generations & 1000 && Population Size & 1024\\
			Mutation & 20\% && Crossover & 80\% \\
			Elitism & top 10 && Selection Type & Tournament\\
			Min. Tree Depth & 2 && Max. Tree Depth & 8\\
			Tournament Size & 7 && Pop. Initialisation & Half-and-half\\
			
			\bottomrule
		\end{tabularx}%
		\vspace{-1em}
	\end{table}
	

	\section{Results and Analysis}
	The full set of results for each method and dataset are shown in Table \ref{results}. For each baseline method on each dataset, we label the result with a ``$+$'' if the baseline was significantly \textbf{better} than GP-MaL, a ``$-$'' if it was significantly \textbf{worse}. If neither of these notations appear, there was no significant difference in the results. We used one-tailed Mann-Whitney $U$ tests with a 95\% confidence interval to compute significance. For convenience, a summary of these results are provided in Table \ref{resultsSummary} by totalling the number of ``wins'' (significantly better), ``losses'' (significantly worse) and ``draws'' (no significant difference) the proposed GP-MaL method has compared with each baseline. We compare GP-MaL's performance to PCA and MDS, and LLE and t-SNE in the following subsections, as these pairs of methods exhibit similar patterns.

	\begin{table}[hbt!]
	\footnotesize
		\vspace{-1em}
		\centering
			\caption{Experiment Results. GPM refers to the proposed GP-MaL method. The number after each method specifies the dimensionality of the low-dimensional manifold; ``cr'' means the cube root approach determined the dimensionality.}
		\label{results}
		\begin{tabular}{@{}llllllllllll@{}}
			\toprule
			Method & Wine & Move. & Derm. & Iono. & Image. & COIL20 & Mad. & Yale & MFAT & MNIST &  \\ \midrule
			GPM2 & 0.955 & 0.485 & 0.914 & 0.826 & 0.797 & 0.628 & 0.605 & 0.382 & 0.639 & 0.909 &  \\
			PCA2 & 0.764$-$ & 0.405$-$ & 0.769$-$ & 0.776$-$ & 0.675$-$ & 0.647$+$ & 0.572$-$ & 0.244$-$ & 0.643 & 0.906$-$ &  \\
			MDS2 & 0.711$-$ & 0.476$-$ & 0.723$-$ & 0.837$+$ & 0.716$-$ & 0.732$+$ & 0.574$-$ & 0.339$-$ & 0.687$+$ & 0.909 &  \\
			LLE2 & 0.659$-$ & 0.499$+$ & 0.803$-$ & 0.833 & 0.809$+$ & 0.850$+$ & 0.601$-$ & 0.120$-$ & 0.843$+$ & 0.980$+$ &  \\
			tSNE2 & 0.718$-$ & 0.782$+$ & 0.852$-$ & 0.890$+$ & 0.921$+$ & 0.948$+$ & 0.712$+$ & 0.455$+$ & 0.935$+$ & 0.986$+$ &  \\ \midrule
			GPM3 & 0.964 & 0.579 & 0.924 & 0.872 & 0.892 & 0.773 & 0.688 & 0.472 & 0.765 & 0.925 &  \\
			PCA3 & 0.793$-$ & 0.608$+$ & 0.780$-$ & 0.877 & 0.805$-$ & 0.823$+$ & 0.681$-$ & 0.374$-$ & 0.749$-$ & 0.932$+$ &  \\
			MDS3 & 0.726$-$ & 0.594$+$ & 0.774$-$ & 0.910$+$ & 0.883$-$ & 0.849$+$ & 0.677$-$ & 0.404$-$ & 0.830$+$ & 0.932$+$ &  \\
			LLE3 & 0.667$-$ & 0.513$-$ & 0.824$-$ & 0.847$-$ & 0.831$-$ & 0.923$+$ & 0.648$-$ & 0.297$-$ & 0.847$+$ & 0.984$+$ &  \\
			tSNE3 & 0.712$-$ & 0.768$+$ & 0.847$-$ & 0.756$-$ & 0.924$+$ & 0.952$+$ & 0.731$+$ & 0.394$-$ & 0.935$+$ & 0.987$+$ &  \\ \midrule
			GPM5 & 0.960 & 0.673 & 0.951 & 0.915 & 0.958 & 0.847 & 0.864 & 0.553 & 0.888 & 0.940 &  \\
			PCA5 & 0.913$-$ & 0.705$+$ & 0.899$-$ & 0.923$+$ & 0.911$-$ & 0.887$+$ & 0.881$+$ & 0.531$-$ & 0.885 & 0.945$+$ &  \\
			MDS5 & 0.732$-$ & 0.719$+$ & 0.817$-$ & 0.928$+$ & 0.901$-$ & 0.886$+$ & 0.685$-$ & 0.564$+$ & 0.881$-$ & 0.948$+$ &  \\
			LLE5 & 0.683$-$ & 0.684$+$ & 0.825$-$ & 0.817$-$ & 0.837$-$ & 0.930$+$ & 0.665$-$ & 0.456$-$ & 0.870$-$ & 0.985$+$ &  \\
			tSNE5 & 0.718$-$ & 0.747$+$ & 0.835$-$ & 0.714$-$ & 0.930$-$ & 0.878$+$ & 0.763$-$ & 0.532$-$ & 0.939$+$ & 0.987$+$ &  \\ \midrule
			GPMcr & 0.962 & 0.681 & 0.941 & 0.899 & 0.891 & 0.913 & 0.863 & 0.661 & 0.935 & 0.952 &  \\
			PCAcr & 0.789$-$ & 0.704$+$ & 0.852$-$ & 0.879$-$ & 0.804$-$ & 0.950$+$ & 0.857$-$ & 0.648$-$ & 0.939$+$ & 0.957$+$ &  \\
			MDScr & 0.725$-$ & 0.722$+$ & 0.792$-$ & 0.920$+$ & 0.884$-$ & 0.911 & 0.670$-$ & 0.651 & 0.889$-$ & 0.957$+$ &  \\
			LLEcr & 0.669$-$ & 0.685 & 0.814$-$ & 0.803$-$ & 0.828$-$ & 0.924$+$ & 0.679$-$ & 0.577$-$ & 0.912$-$ & 0.984$+$ &  \\
			tSNEcr & 0.710$-$ & 0.759$+$ & 0.853$-$ & 0.713$-$ & 0.925$+$ & 0.730$-$ & 0.765$-$ & 0.650$-$ & 0.944$+$ & 0.987$+$ &  \\ \bottomrule
		\end{tabular}

	\caption{Summary of Experiment Results. The number of ``wins'', ``losses'', and ``draws'' are shown for GP-MaL compared to each baseline.}
	\label{resultsSummary}
	\begin{tabularx}{0.75\textwidth}{@{}ccccXcccc@{}}
		\toprule
Baseline & Wins & Losses & Draws &  & Baseline & Wins & Losses & Draws \\ 
\cmidrule{1-4}\cmidrule{6-9}
PCA2 & 8 & 1 & 1 &  & PCA3 & 6 & 3 & 1 \\
MDS2 & 6 & 3 & 1 &  & MDS3 & 5 & 5 & 0 \\
LLE2 & 4 & 5 & 1 &  & LLE3 & 7 & 3 & 0 \\
tSNE2 & 2 & 8 & 0 &  & tSNE3 & 4 & 6 & 0 \\ \cmidrule{1-4}\cmidrule{6-9}
PCA5 & 4 & 5 & 1 &  & PCAc & 6 & 4 & 0 \\
MDS5 & 5 & 5 & 0 &  & MDSc & 5 & 3 & 2 \\
LLE5 & 7 & 3 & 0 &  & LLEc & 7 & 2 & 1 \\
tSNE5 & 6 & 4 & 0 &  & tSNEc & 6 & 4 & 0 \\ \bottomrule
	\end{tabularx}
\vspace{-2em}
\end{table}
\subsection{GP-MaL compared to PCA \& MDS}
GP-MaL has a clear advantage over PCA when the most significant amount of feature reduction --- to two or three features --- is required. Given that PCA is a linear manifold learning method, it is not surprising that GP-MaL is able to preserve more structure in 2 or 3 dimensions by performing more complex, non-linear reductions. At higher dimensions, the gap narrows somewhat, as at 5 or CR features there are enough available output dimensions in order to make linear combinations able to model the underlying structure of the data more accurately. PCA weights every input feature in each component it creates, which means the way in which it models this structure is rather opaque when there are many input features. The MDS results have a similar pattern to the PCA ones, except that MDS and GP-MaL are quite even on 3 and 5 features. It is interesting to note that MDS uses a similar optimisation criterion to GP-MaL, but struggles significantly more at 2 features.

\subsection{GP-MaL compared to LLE \& t-SNE}
Overall, GP-MaL is the most consistent of all the methods across the different numbers of features produced. LLE wins on one more dataset than GP-MaL for 2 features, but otherwise GP-MaL has a clear advantage with 7 wins on 3/5/CR features. The performance of LLE fluctuates quite widely across the datasets, and generally loses to PCA as the number of features is increased.

While GP-MaL is clearly worse than t-SNE on the 2 and 3D results, it outperforms t-SNE on 5 or CR features. On the Ionosphere and COIL20 datasets, t-SNE's performance actually decreases as the number of output features are increased, which means it is much more sensitive to the number of components that the user chooses than GP-MaL; GP-MaL almost strictly improves as more output features are produced, which is what we generally expect from dimensionality reduction techniques. 

In a number of cases, t-SNE does actually outperform GP-MaL while using fewer features --- however, consider that t-SNE (and LLE) are embedding method which do not have to manipulate the original feature space to produce the output feature space (i.e. it is not a functional mapping). It is clearly more difficult to evolve such a mapping, but also has significant benefits in that GP-MaL's output dimensions can be interpreted in terms of how they combine the original features, which is often as important as visualisation alone in exploratory data analysis. This behaviour will be explored further in Section \ref{furtherAnalysis}.

\subsection{Summary}
GP-MaL shows promising performance for an initial attempt at directly using GP for manifold learning, winning against all baselines on at least two of the four configurations tested. While GP-MaL faltered somewhat on some datasets such as MNIST, it achieved much better performance on other lower-dimensional datasets such as Wine and Dermatology. This suggests that with further improvements to its learning capacity, GP-MaL may have the potential to outperform existing methods on these higher-dimensional datasets too.

Another important consideration is the interpretability of the models produced by each baseline. t-SNE, LLE, MDS, and PCA (to a lesser extent) are almost black-boxes as they give little information about how the manifolds in the data are represented in terms of the original features. Interpretability is an increasing concern in data mining, and feature reduction is often touted as a way to improve it; we will examine in the following section if GP-MaL can be interpreted any more easily than these existing methods.

	\section{Further Analysis}
	\label{furtherAnalysis}
	\subsection{GP-MaL for Data Visualisation}

	\def \scaleFactor{0.33}
	\def \path{figures/}
	\def \spacey{\hspace{-.7em}}
\begin{figure}[p]
	\vspace{-2em}
	\centering
		\subfloat[GP-MaL]{
			\includegraphics[width=\scaleFactor\textwidth]{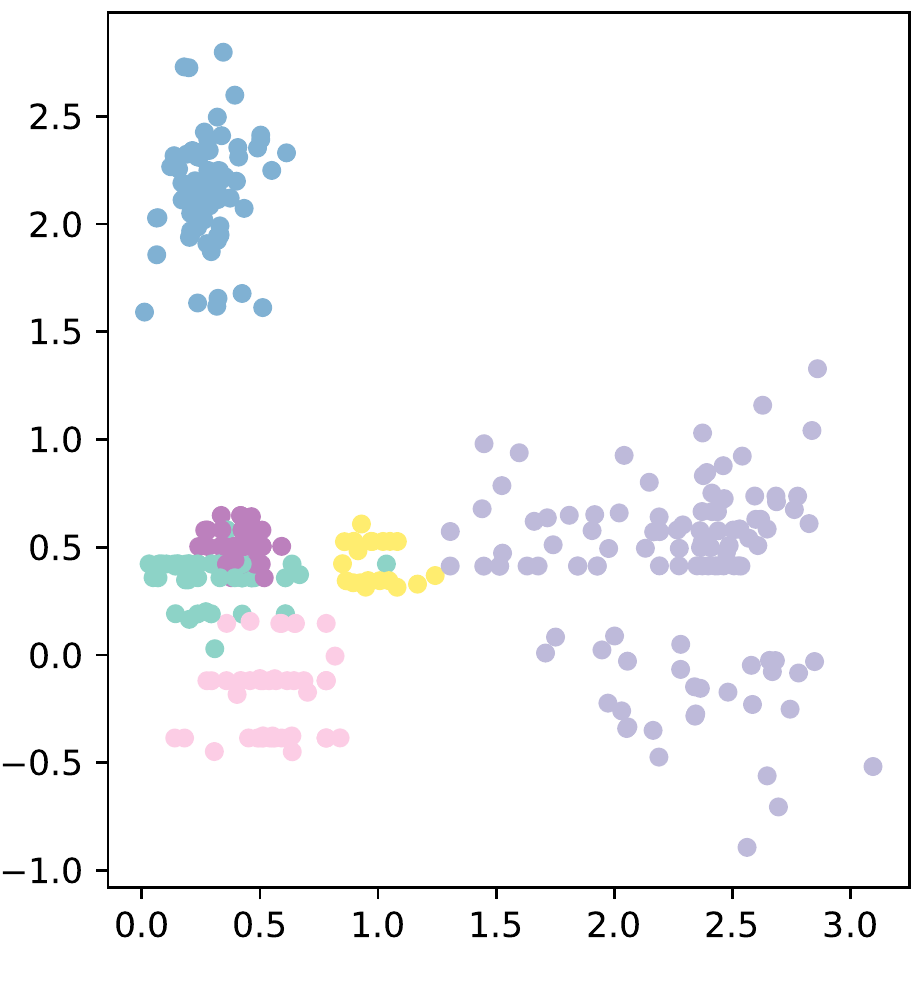}			\spacey
			\label{dermGP}
		} 
	\subfloat[PCA]{
		\label{dermPCA}
\includegraphics[width=\scaleFactor\textwidth]{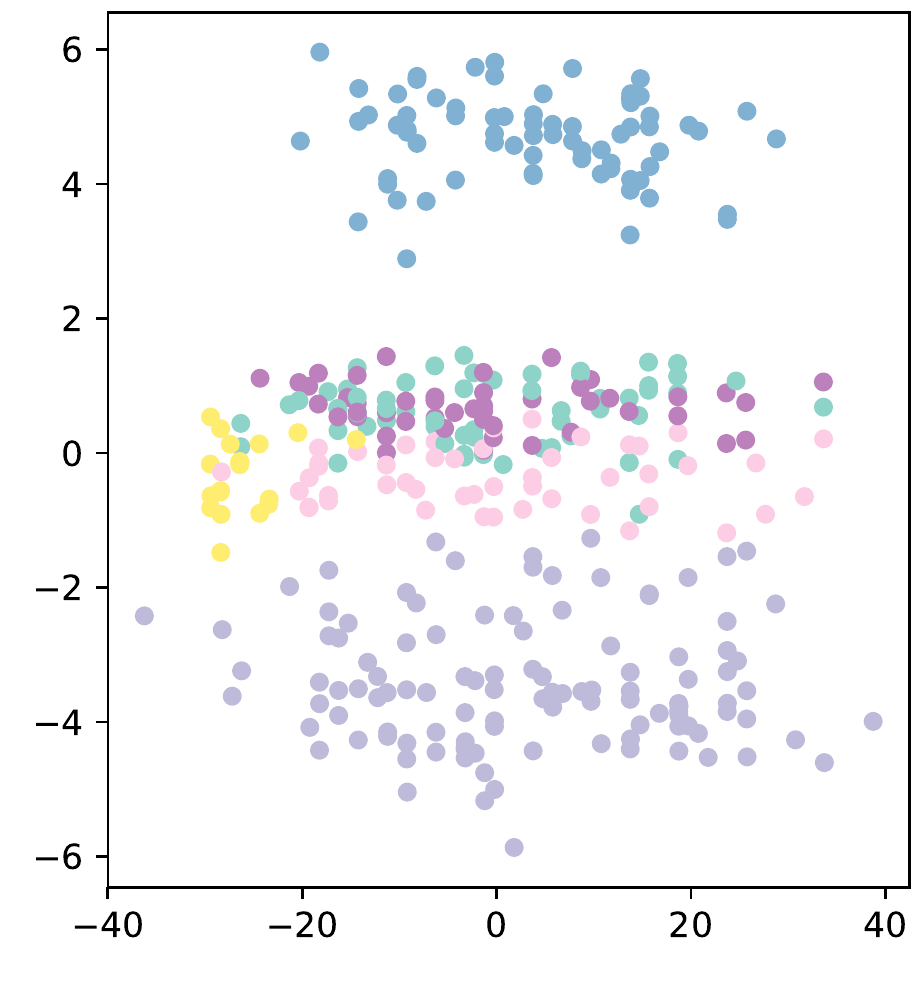}
		\spacey
	} 
\subfloat[MDS]{
	\label{dermMDS}
\includegraphics[width=\scaleFactor\textwidth]{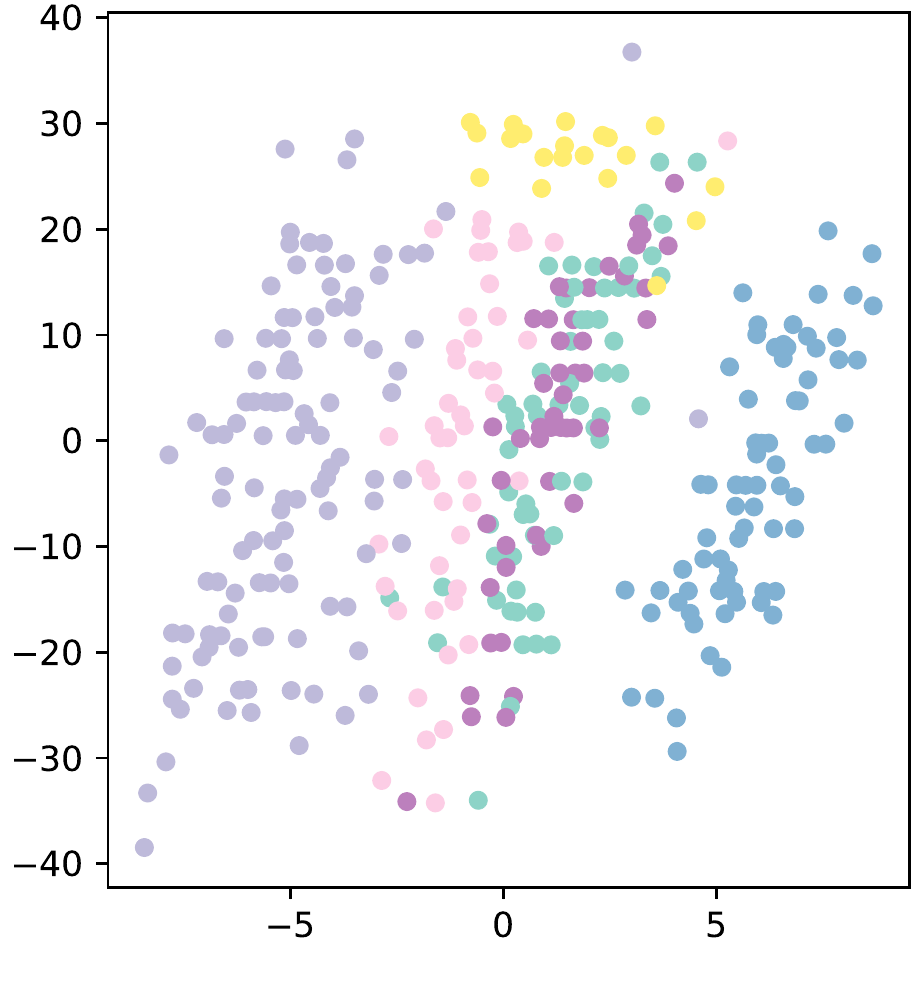}
	\spacey
} 
\vspace{-1em}
\subfloat[LLE]{
	\label{dermLLE}
\includegraphics[width=\scaleFactor\textwidth]{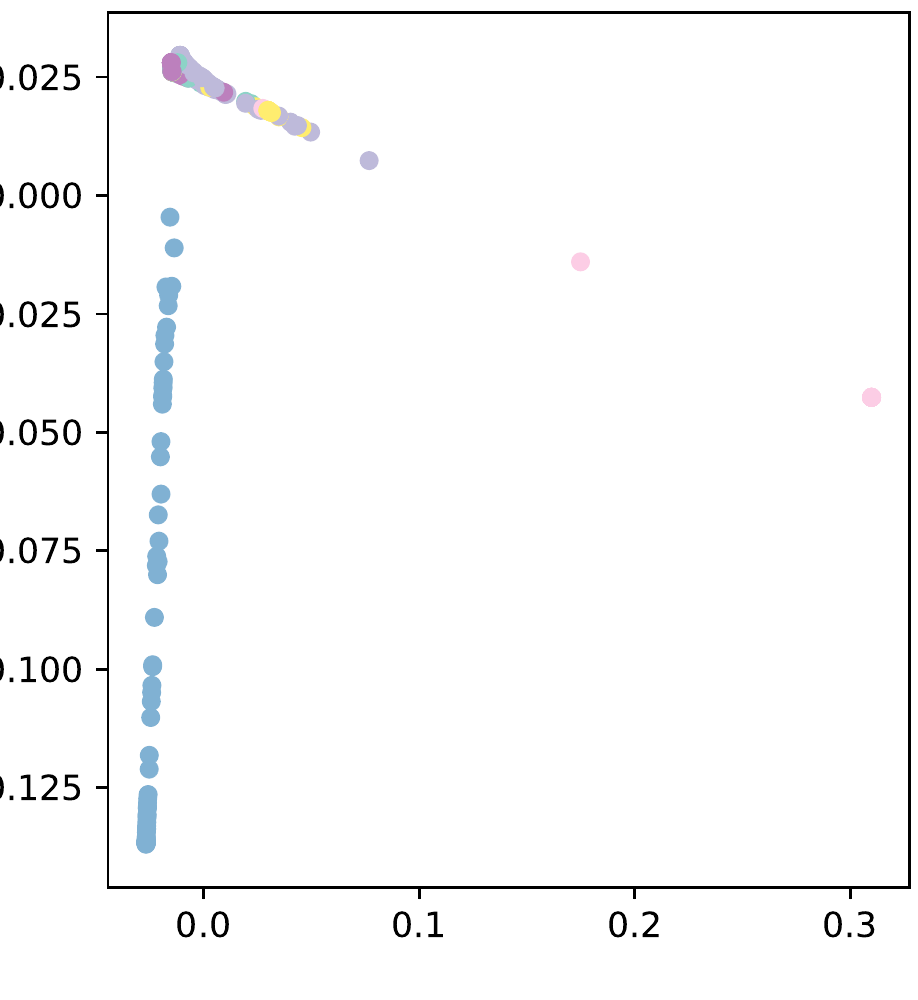}
	\spacey
} 
\subfloat[tSNE]{
	\label{dermtSNE}
\includegraphics[width=\scaleFactor\textwidth]{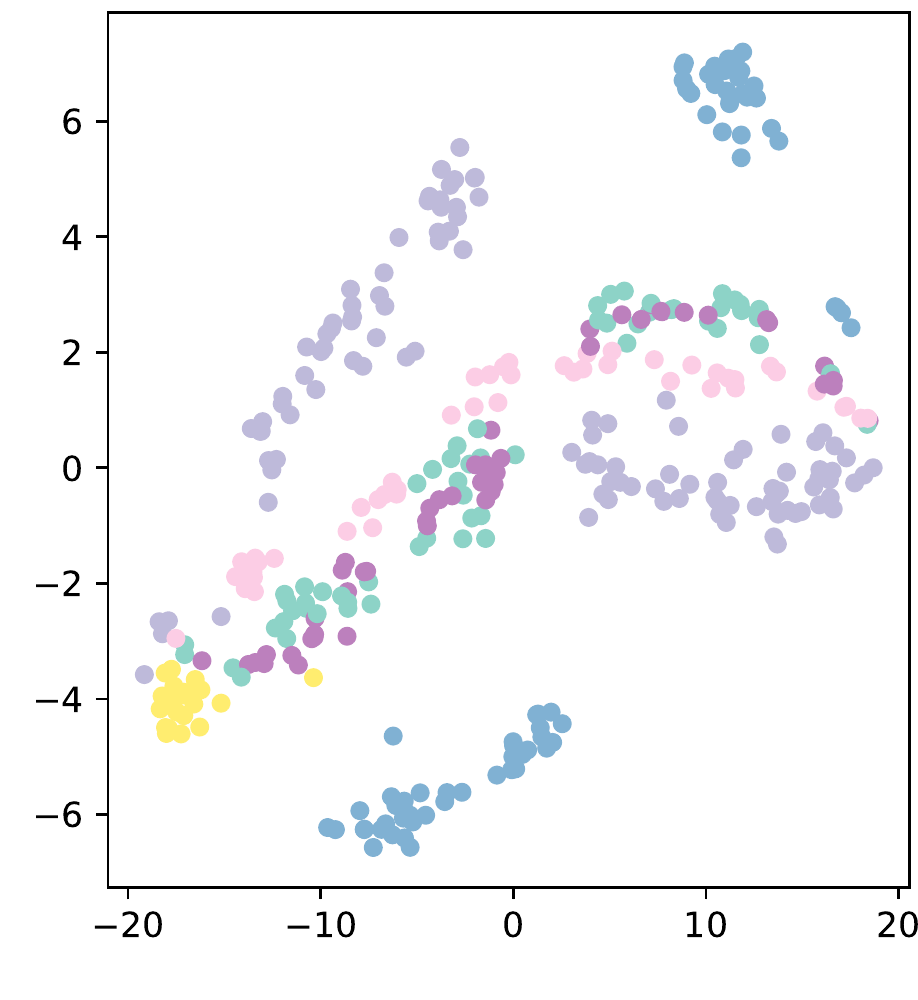}
	\spacey
} 

\caption{The two created features on Dermatology dataset, coloured by class label.}
	\label{dermVis}
		\centering
	\subfloat[GP-MaL]{
		\includegraphics[width=\scaleFactor\textwidth]{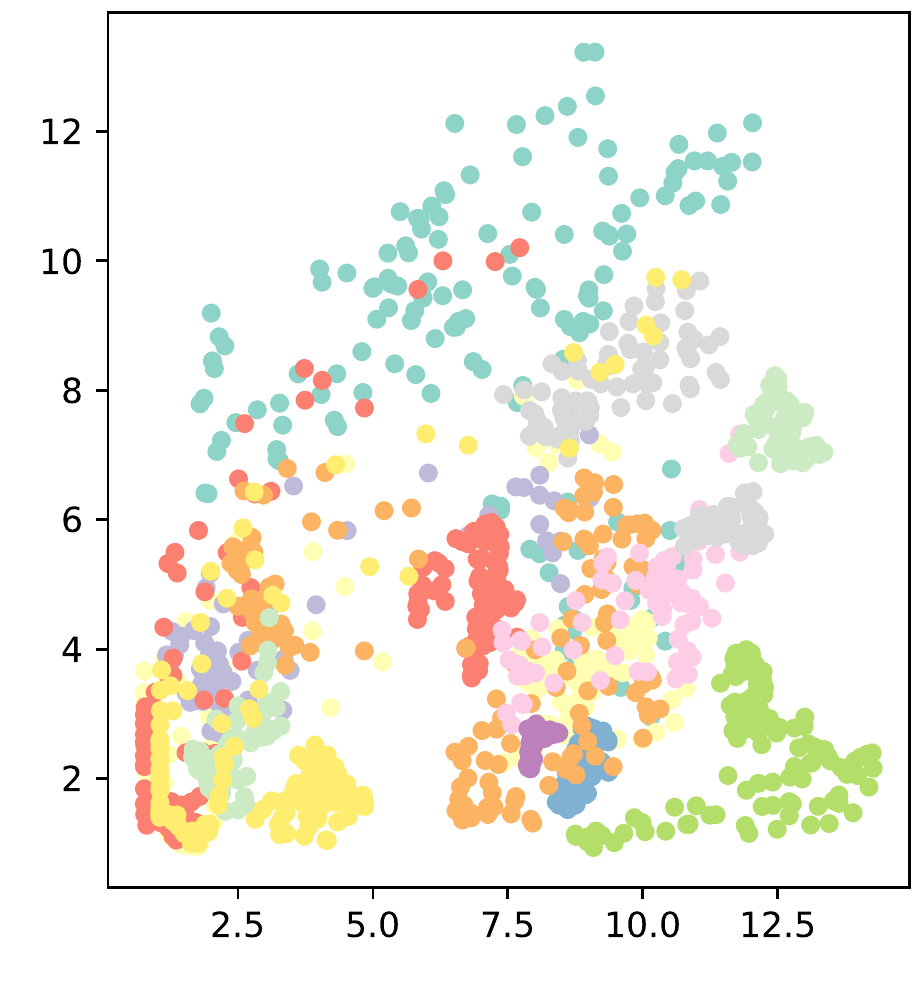}			\spacey
		\label{coil20GP}
	} 
	\subfloat[PCA]{
		\label{coil20PCA}
		\includegraphics[width=\scaleFactor\textwidth]{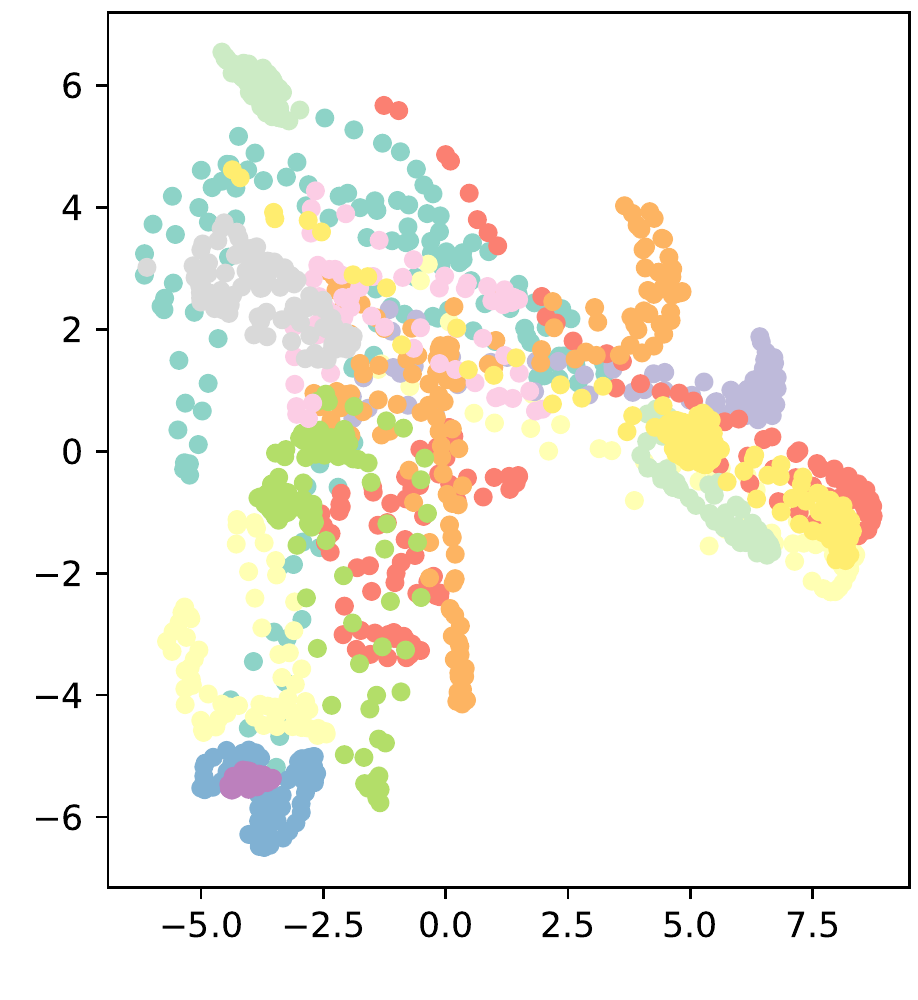}
		\spacey
	} 
	\subfloat[MDS]{
		\label{coil20MDS}
		\includegraphics[width=\scaleFactor\textwidth]{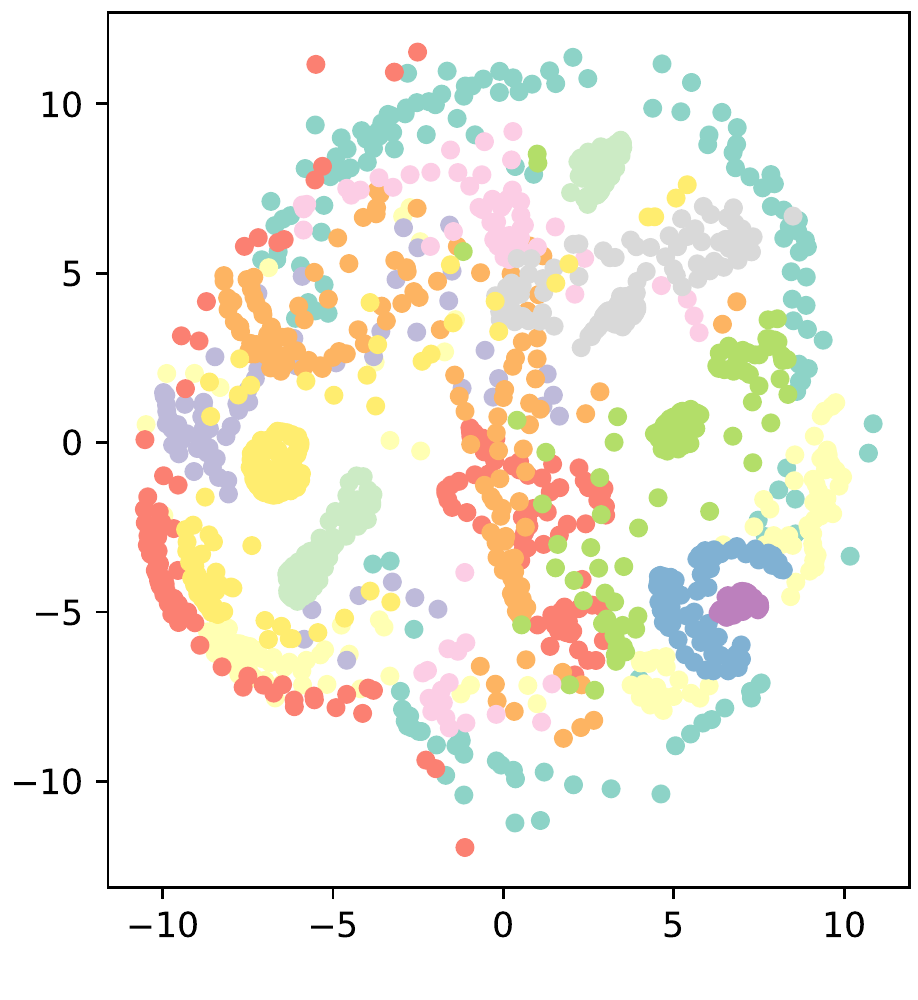}
		\spacey
	} 
	\vspace{-1em}
	\subfloat[LLE]{
		\label{coil20LLE}
		\includegraphics[width=\scaleFactor\textwidth]{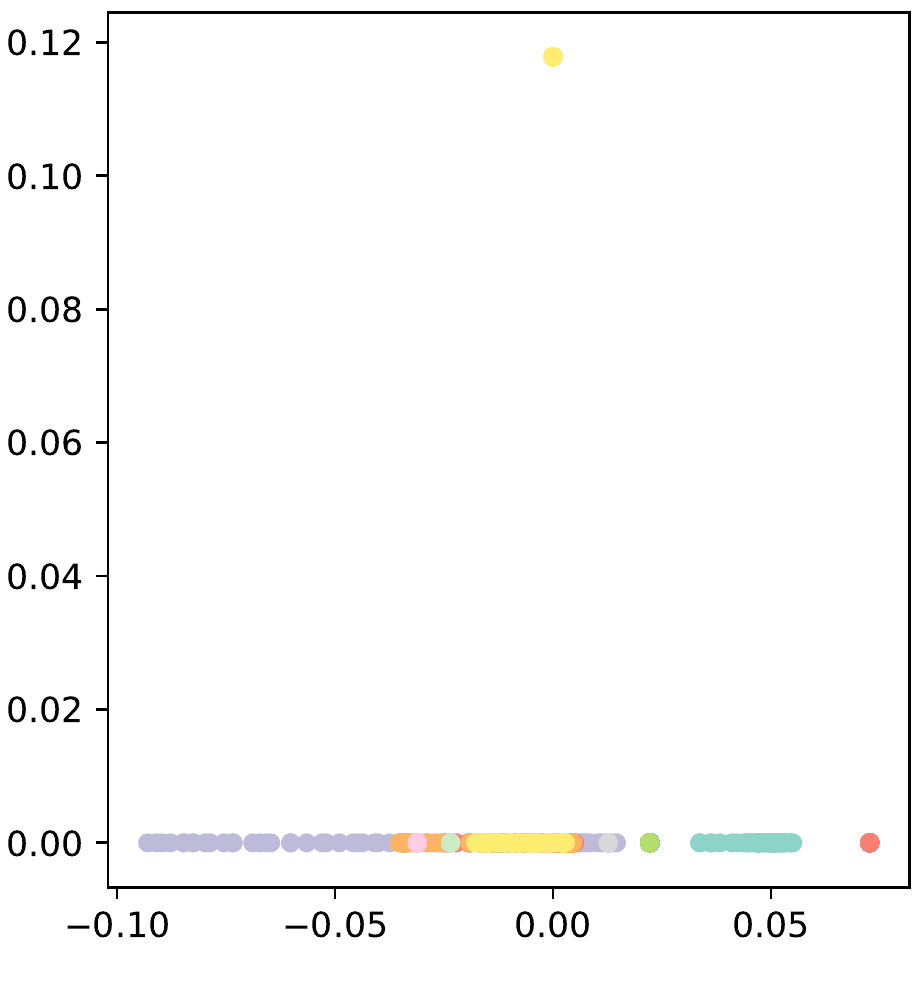}
		\spacey
	} 
	\subfloat[tSNE]{
		\label{coil20tSNE}
		\includegraphics[width=\scaleFactor\textwidth]{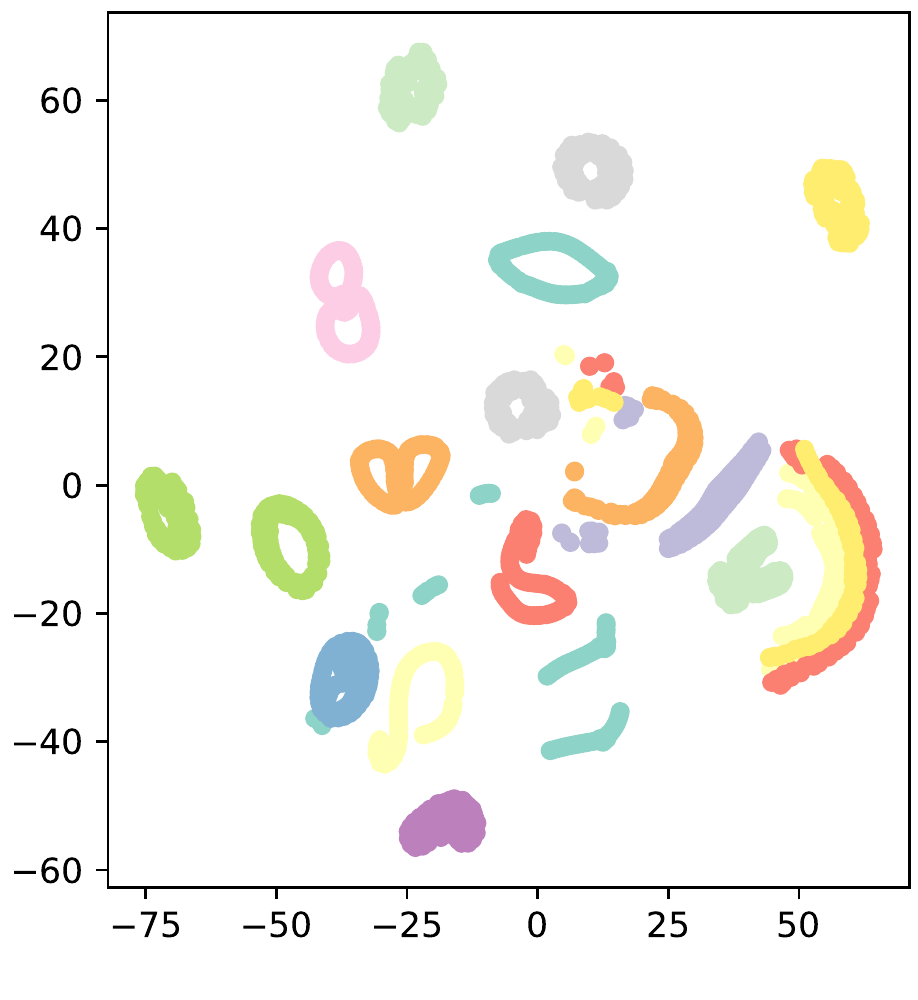}
		\spacey
	} 
	
	\caption{The two created features on COIL20 dataset, coloured by class label.}
	\label{coilVis}
\end{figure}

A common use of manifold learning techniques such as t-SNE and PCA is for visualisation of datasets in two- or three-dimensions. Figs.\ \ref{dermVis} and \ref{coilVis} plot the two-dimensional outputs of each manifold learning method for two datasets that GP-MaL performed best on (Dermatology) and worst on (COIL20) respectively. To show the potential of each method, we used the result that had the highest classification accuracy for plotting.

On the Dermatology dataset, GP-MaL clearly separates each class better than the baseline manifold methods. PCA, MDS, and t-SNE struggle to seperate the purple, green, and pink classes apart, whereas GP-MaL is able to separate them while keeping them reasonably close to signify their similarities. t-SNE splits both the purple and blue classes into two disjoint groups with other classes appearing in the middle of the split. LLE clearly struggles to give a good visualisation at all --- it is only able to split the blue class from the others. 

\begin{figure}[hbt!]

\end{figure}

On the COIL20 dataset, LLE is able to separate the classes somewhat more effectively along one dimension, but still fails to produce a reasonable visualisation. t-SNE clearly does very well, but does continue to separate some classes into disjoint clusters (all of the green ones). It is not clear which of GP-MaL, PCA, and MDS produces the best result; MDS tends to incorrectly separate some classes like t-SNE, whereas GP-MaL and PCA have poorer separation of different classes overall. Unlike the other methods, the two dimensions produced by GP-MaL can be interpreted in terms of how they use the original features --- this will be explored further in the next subsection.

\subsection{Tree Interpretability}
\begin{figure}[hbt!]
		\vspace{-1.5em}
		\centering
		
		\subfloat[]{\includegraphics[width=0.85\linewidth]{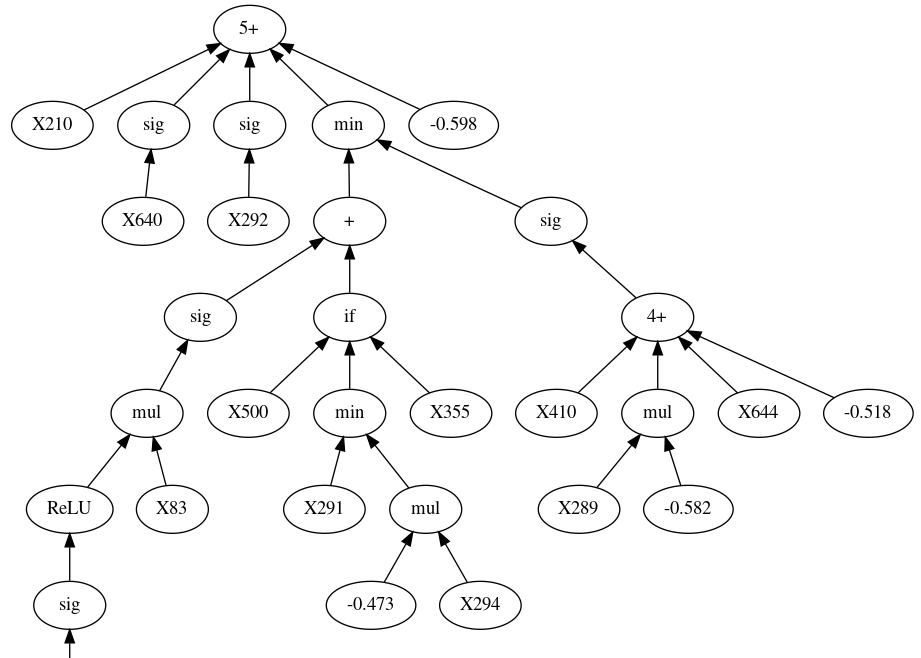}}
				\hspace{-6em}\subfloat[]{\boxed{\includegraphics[width=0.2\linewidth]{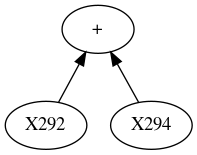}}}
		\caption{An example of two simplified trees (features) evolved on the MFAT dataset, giving 65\% classification accuracy. Only the top of the left tree is shown.}
		\label{fig:tree1mfatpartial}
		\vspace{-1.5em}

	\end{figure}

Part of an individual evolved on the MFAT dataset is shown in Fig.\ \ref{fig:tree1mfatpartial}. While the left tree is quite large (containing 73 features), the right tree is very simple: it adds two features together, and outputs the sum. This gives a strong indication that these features are very important for modelling the underlying structure of the data. While the left tree is harder to analyse, it is useful to note that four of the children of the root node are again very simple: three features (two of which are transformed non-linearly) and a constant weighting. This again suggests that these features particularly model the instances in the MFAT dataset, with $X292$ appearing in both trees. $X292$ and $X294$ are the first and third Karhunen-Lo\`eve coefficients extracted from the original images; these coefficients are extracted in a similar way to PCA, so it makes sense that GP would recognise these to be very useful features: being the first and third coefficients, they represent a significant amount of the variance present in this dataset. 

		\begin{figure}[hbt!]
		
		\centering
		\vspace{-1.5em}

				\subfloat[]{\includegraphics[width=0.85\linewidth]{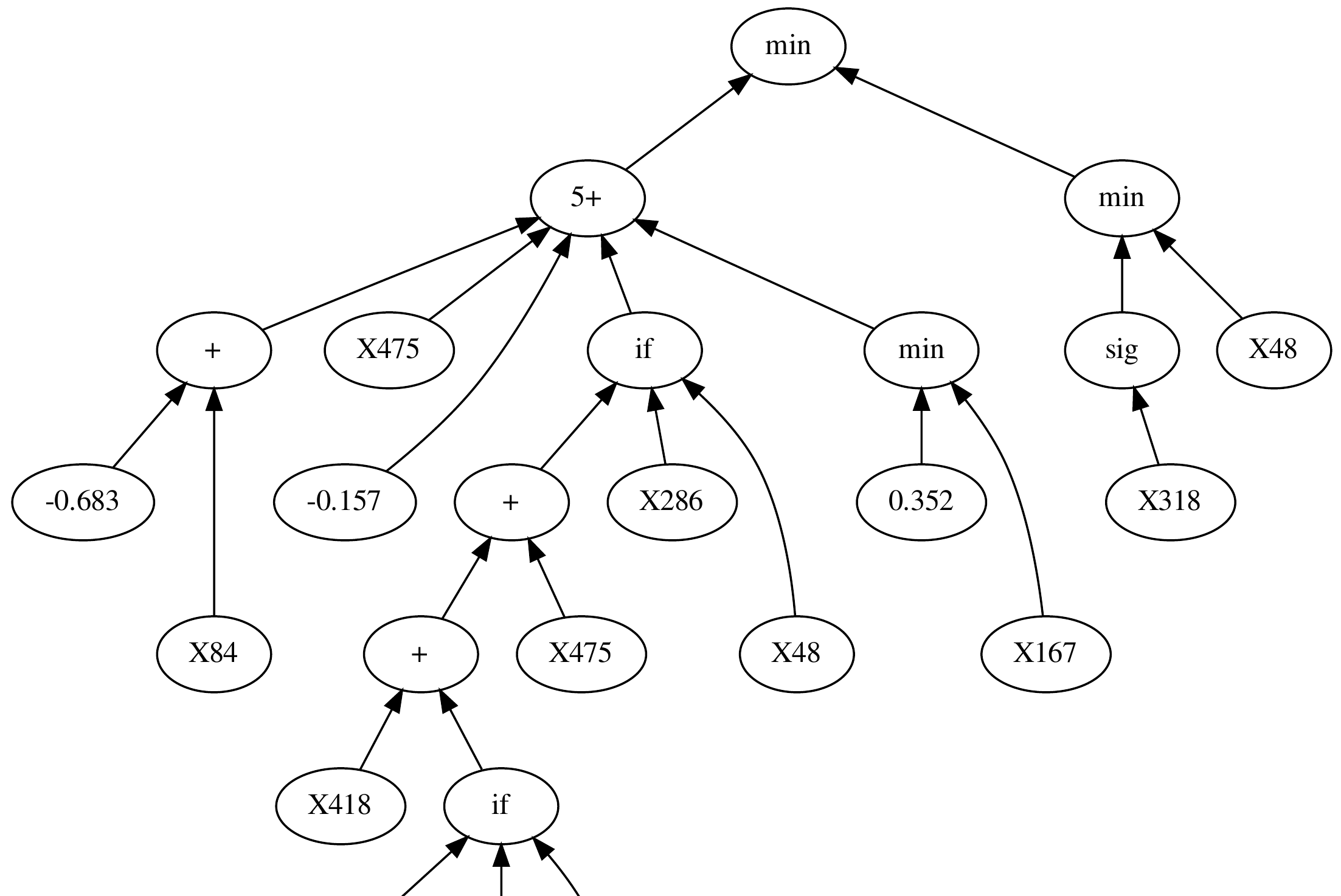}\hspace{-2em}\label{madelon:tree0}}
						\subfloat{\boxed{\includegraphics[width=0.2\linewidth]{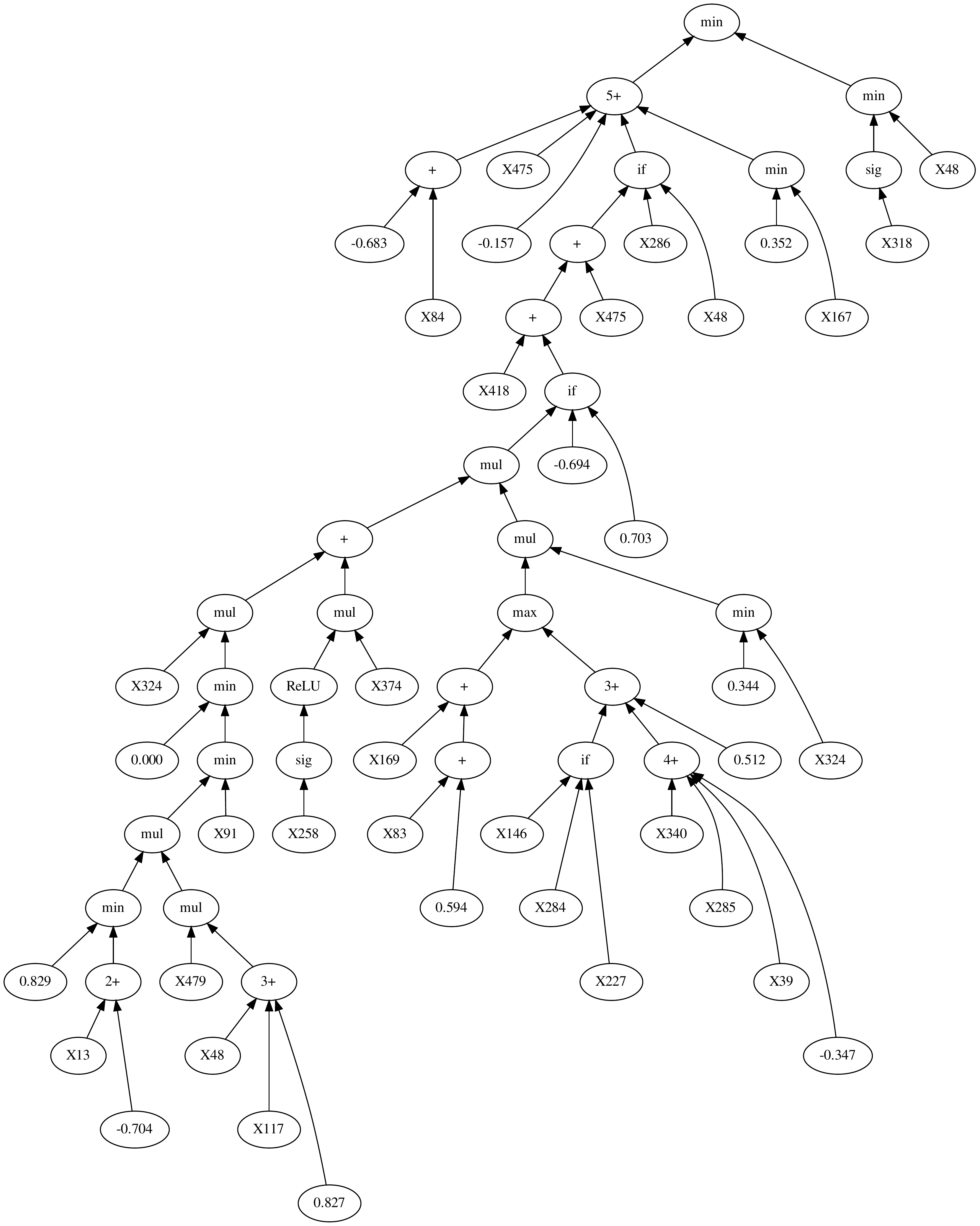}}}
		\vspace{-1.5em}
		\setcounter{subfigure}{1}
						\hspace{-.2em}
		\subfloat[]{\includegraphics[width=0.26\linewidth]{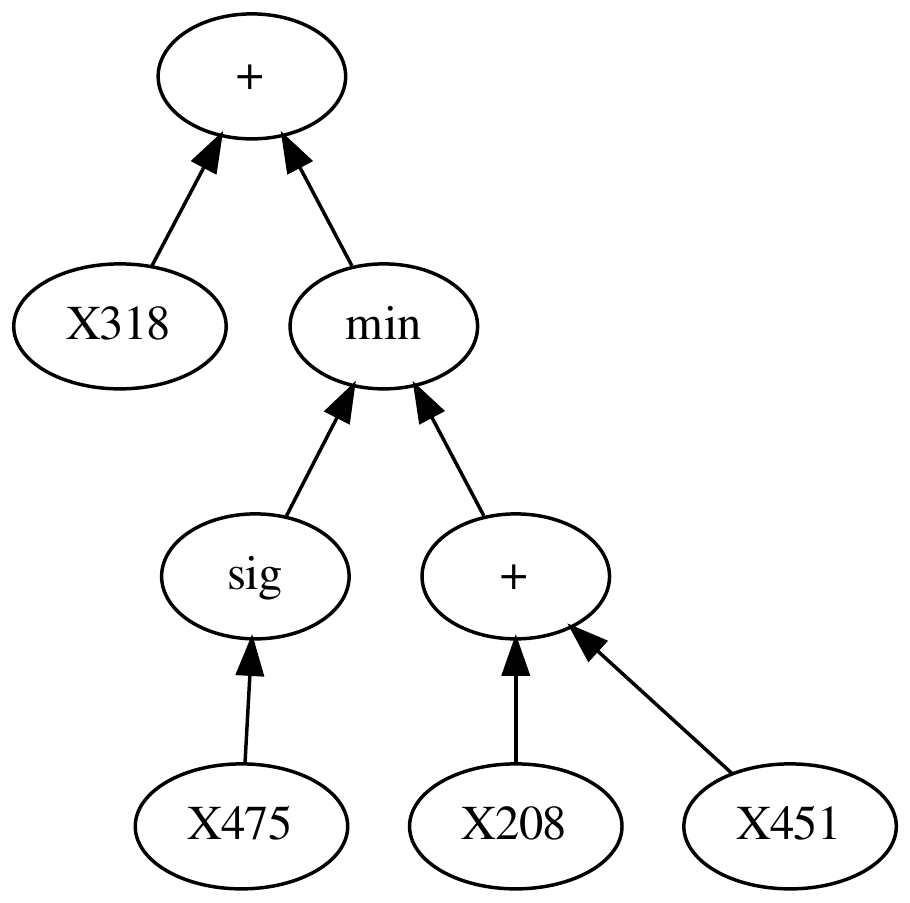}\label{madelon:tree1}}
				\hspace{-.4em}
		\subfloat[]{\includegraphics[width=0.22\linewidth]{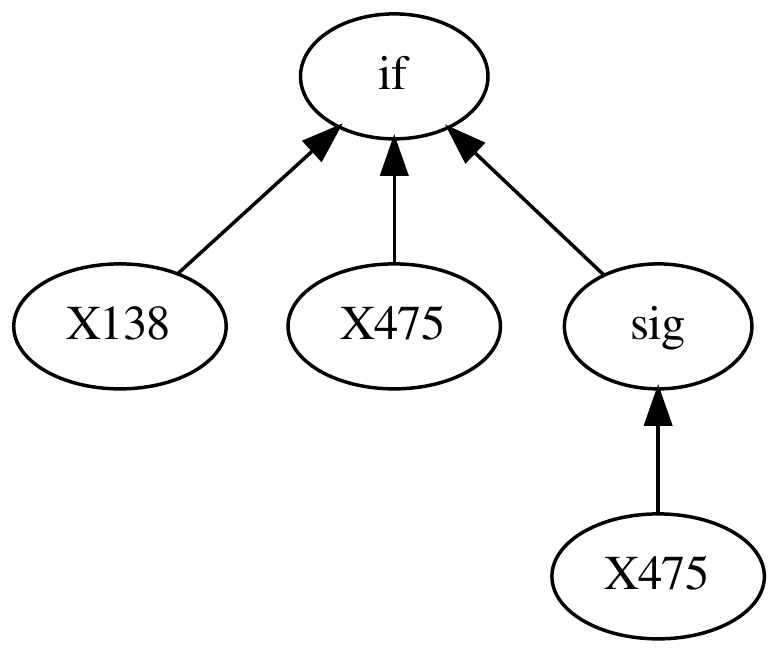}\label{madelon:tree2}}
						\hspace{.4em}
		\subfloat[]{\includegraphics[width=0.45\linewidth]{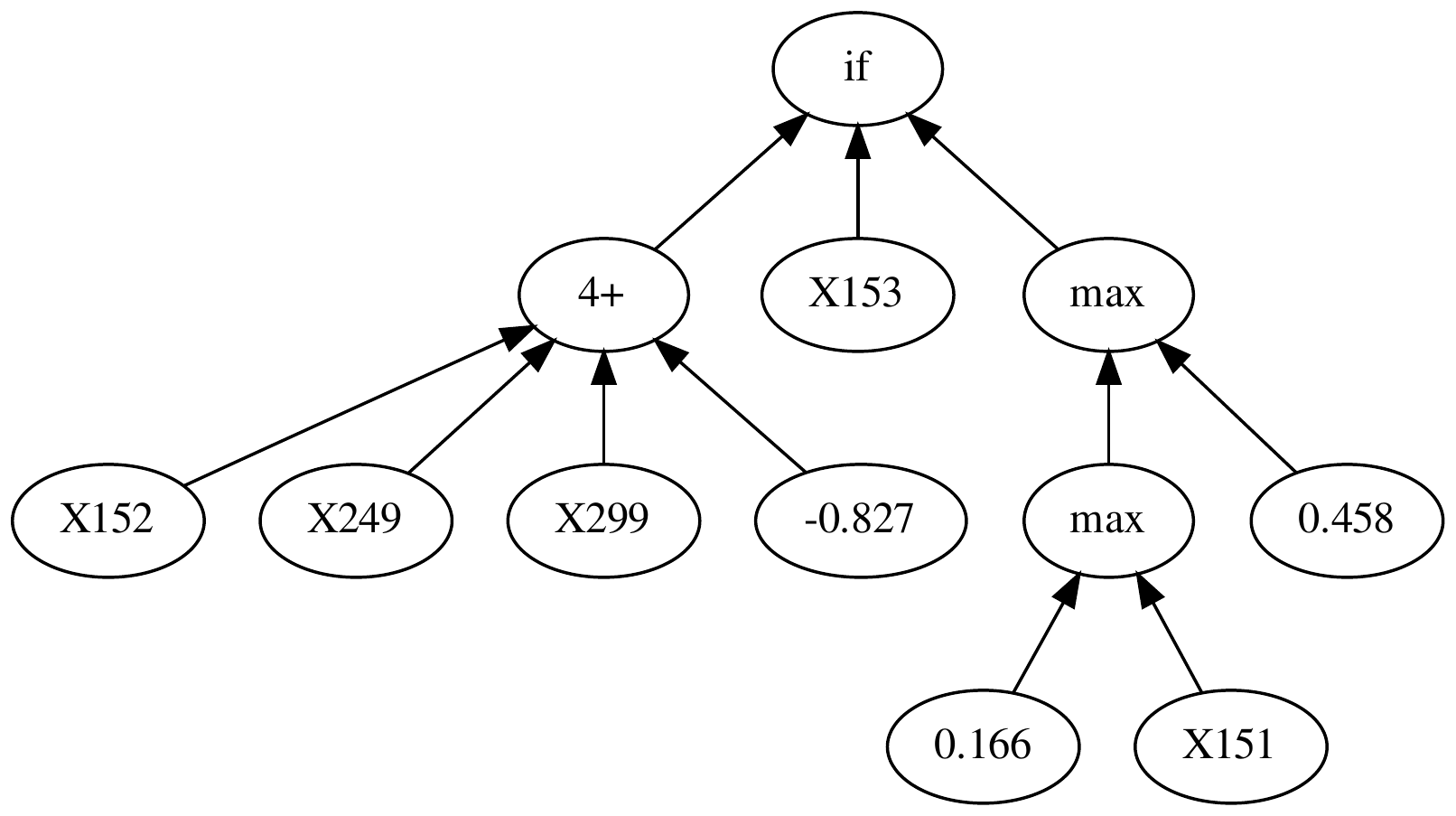}\label{madelon:tree3}}
		\hspace{-1em}
		\subfloat[]{\includegraphics[width=0.07\linewidth]{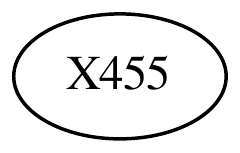}\label{madelon:tree4}}
		\caption{An example of five simplified trees (features) evolved on the Madelon dataset, giving 87.8\% classification accuracy. Tree (a) was truncated to save space, but the full tree in the box to the right to show that is is reasonably small. The original dataset had 500 features.}
		\label{fig:madelonTree}
		\vspace{-2em}
	\end{figure}
	
	Fig.\ \ref{fig:madelonTree} shows another example GP individual evolved on the 500-dimensional Madelon dataset, with five trees used. Of the five trees, four are simple enough to be human-interpretable, with the fifth being somewhat larger, but still interpretable at the root. Trees \ref{madelon:tree1} -- \ref{madelon:tree4} each combine between three and five features in ways that make sense, but which would not be able to be represented by many existing manifold learning methods. For example, consider Tree \ref{madelon:tree2}, which uses either the original value of $X475$, or a non-linear sigmoid transformation of $X475$ depending on the value of $X138$. This suggests that there is a particular feature interaction between $X138$ and $X475$ which may be important to the underlying structure of the dataset. In fact, $X475$ is the feature which has the second-highest information gain (IG) on this dataset\footnote{Information gain (mutual information) is often used in feature selection for classification to measure the dependency between a feature and the class label.}. Tree \ref{madelon:tree4} is just a single selected feature --- $X455$ clearly is important in the manifold of this dataset.
	
	Although Tree \ref{madelon:tree0} is clearly more complex, the top of the tree still provides an interesting picture of the most important aspects of the dataset. For example, if $X48$ is a very low value, then this is simply the output of the tree. Examining $X48$ more closely reveals that it is in the top 3\% of features in terms of IG, and that at its smallest values it always predicts the positive class. Also of note is that $X475$ appears twice again in the top of this tree as well as in Tree \ref{madelon:tree2}. 
	
	\subsubsection{Summary:}
	As the focus of this work was to show the \textit{potential} of GP for direct manifold learning, no parsimony pressure (or other such methods) were applied to encourage simple trees. Nevertheless, aspects of the evolved individuals can be analysed with ease and provide insight into the structure of the datasets. This is a clear advantage over existing manifold learning techniques which are black (or very grey) boxes, and bodes well for future work. The use of GP also has the nice upside of allowing the evolved trees to be re-used on future examples without having to perform the whole manifold learning process again (as t-SNE or the other methods would require).

	\section{Conclusion}
Manifold learning has become significantly more popular in recent years due to the emergence of autoencoders and visualisation techniques such as t-SNE. Despite this, the learnt manifolds tend to be completely or nearly uninterpretable with respect to the original feature space. With interpretability being a key goal in exploratory data mining, we proposed a new GP method called GP-MaL for directly performing manifold learning in this paper. Appropriate terminal and function sets were presented, along with a fitness function tailored to the task, and further techniques for reducing computational complexity. We showed that GP-MaL was competitive, and in some cases clearly better than, existing manifold learning methods, and was generally more stable across that datasets tested. GP-MaL was also shown to produce interpretable models that help the user to gain concrete insight into their dataset, unlike many existing manifold learning methods. The potential of GP-MaL for visualisation purposes was highlighted. Furthermore, the functional nature of the models produced by GP-MaL allows it to be applied to future data without re-training. As the first work using GP for directly performing manifold learning, these findings showcase the potential of GP to be applied further to this domain.

GP-MaL is quite flexible, and could easily be extended further with other function sets and fitness functions (that do not have to be differentiable!). We also hope to explore a multi-objective approach in the future that balances the often-conflicting objectives of maintaining global and local structure. It is also clear that techniques to encourage simpler/more concise trees such as parsimony pressure would further improve the usefulness of GP-MaL.

	
	%
	%
	%
	\bibliographystyle{splncs04}
	%

	\bibliography{lensen}

\end{document}